\documentclass[10pt,twocolumn,letterpaper]{article}

\usepackage[pagenumbers]{iccv} 

\usepackage{amsmath}
\usepackage{url}            
\usepackage{booktabs}       
\usepackage{amsfonts}       
\usepackage{nicefrac}       
\usepackage{microtype}      
\usepackage{xcolor}         
\usepackage[inline, shortlabels]{enumitem}
\usepackage{lipsum}
\usepackage{bm}
\usepackage{algpseudocode}
\usepackage{algorithm}
\usepackage{multirow}
\usepackage{pifont}
\newcommand{\cmark}{\ding{51}}%
\newcommand{\xmark}{\ding{55}}%
\usepackage{colortbl}
\usepackage{graphicx}
\usepackage{overpic}
\usepackage{cuted}
\usepackage{makecell}
\definecolor{dkgreen}{rgb}{0,0.6,0}
\definecolor{gray}{gray}{0.95}
\definecolor{mauve}{rgb}{0.58,0,0.82}

\definecolor{iccvblue}{rgb}{0.21,0.49,0.74}
\usepackage[pagebackref,breaklinks,colorlinks,allcolors=iccvblue]{hyperref}

\def\paperID{2787} 
\def\confName{ICCV}
\def\confYear{2025}

\title{Gaussian Variation Field Diffusion for High-fidelity Video-to-4D Synthesis\vspace{-5mm}}

\author{
    Bowen Zhang$^{1*}$ \quad Sicheng Xu$^{2}$ \quad Chuxin Wang$^{1}$ \quad Jiaolong Yang$^{2}$\\
    Feng Zhao$^{1\dagger}$\quad Dong Chen$^{2\dagger}$ \quad Baining Guo$^{2}$\\
	$^1${University of Science and Technology of China} \quad $^2${Microsoft Research Asia}\\
}

\begin{document}
\maketitle

{
	\renewcommand{\thefootnote}%
	{\fnsymbol{footnote}}
	\footnotetext[1]{Intern at Microsoft Research Asia. $^{\dagger}$Corresponding authors.}
}

\begin{strip}
	\vspace{-55pt}
	\centering
	\includegraphics[width=1.0\textwidth]{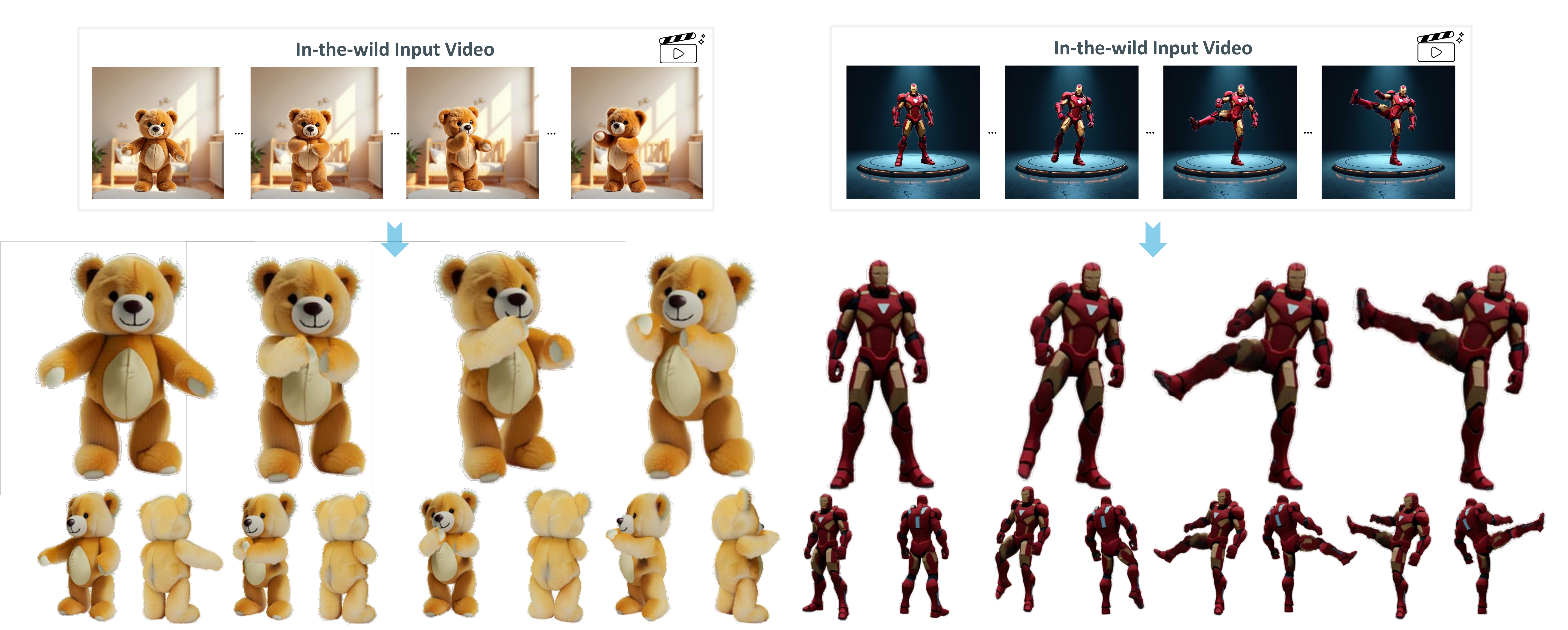}
    \vspace{-10pt}
    \captionsetup{type=figure,font=small,position=top}
    \caption{Our model is capable of creating high-fidelity 4D objects from in-the-wild video inputs. \textbf{\emph{Best viewed with zoom-in.}}} 
    \label{fig:teaser}
    \vspace{-15pt}
\end{strip}

\begin{abstract}
In this paper, we present a novel framework for video-to-4D generation that creates high-quality dynamic 3D content from single video inputs. Direct 4D diffusion modeling is extremely challenging due to costly data construction and the high-dimensional nature of jointly representing 3D shape, appearance, and motion. We address these challenges by introducing a \emph{Direct 4DMesh-to-GS Variation Field VAE} that directly encodes canonical Gaussian Splats (GS) and their temporal variations from 3D animation data without per-instance fitting, and compresses high-dimensional animations into a compact latent space. Building upon this efficient representation, we train a \emph{Gaussian Variation Field diffusion model} with temporal-aware Diffusion Transformer conditioned on input videos and canonical GS. Trained on carefully-curated animatable 3D objects from the Objaverse dataset, our model demonstrates superior generation quality compared to existing methods. It also exhibits remarkable generalization to in-the-wild video inputs despite being trained exclusively on synthetic data, paving the way for generating high-quality animated 3D content. Project page: \href{https://gvfdiffusion.github.io/}{GVFDiffusion.github.io}.
\end{abstract}

\section{Introduction}

Recent advances in generative models have demonstrated remarkable capabilities across various modalities, including image~\cite{karras2019style,karras2020analyzing,zhang2022styleswin,nichol2021improved,ho2022imagen,ramesh2022hierarchical,hang2023efficient}, video~\cite{blattmann2023stable,blattmann2023align,guo2023animatediff}, and 3D content~\cite{tang2023dreamgaussian,zhang2024clay,xiang2024structured,zhang2024gaussiancube,tang2024lgm,hong2023lrm}. While these achievements mark important milestones, they naturally lead to the next frontier: 4D generation, which aims to create dynamic 3D content. The challenge of generating such content—a fundamental aspect of representing our inherently four-dimensional world—remains largely unexplored. This gap is particularly significant given that real-world phenomena inherently combine spatial and temporal dynamics, from the subtle object movements to complex character articulations.

Despite diffusion models~\cite{dhariwal2021diffusion,nichol2021improved,rombach2022high} having demonstrated strong modeling capabilities in both 2D and 3D domains, training a robust 4D diffusion model for dynamic 3D content generation presents two main technical challenges. First, obtaining a large-scale 4D dataset is time-consuming. A straightforward approach involves fitting individual dynamic Gaussian Splatting (4DGS) representations~\cite{wu20244d} for each 3D animation sequence, but this solution typically requires tens of minutes per instance, making it computationally expensive and less scalable as the number of instances increases. Second, the higher-dimensional nature of the problem necessitates a large number of parameters (usually exceeding 100K tokens) to represent 3D shape, appearance, and motion simultaneously, making direct modeling with diffusion approaches extremely challenging. These limitations have significantly hindered the development of efficient and high-quality 4D generative models.

Motivated by the effectiveness of diffusion models applied to compact latent spaces in recent 2D and 3D generation works~\cite{rombach2022high,blattmann2023align, xiang2024structured,zhang2024clay,ren2024xcube}, we present a novel framework for 4D generative modeling that comprises a \emph{Direct 4DMesh-to-GS Variation Field VAE} and a \emph{Gaussian Variation Field diffusion model}. Our VAE framework encodes the canonical 3D Gaussian Splatting (3DGS) of objects and compresses each Gaussian's attribute variations (\ie, \emph{Gaussian Variation Fields}) into a compact latent space from 4D mesh data, thereby bypassing costly per-instance reconstructions. Inspired by previous works~\cite{zhang20233dshape2vecset, cao2024motion2vecsets}, we employ a perceiver-style transformer network~\cite{jaegle2021perceiver,jaegle2021perceiverio,vaswani2017attention} with displacements of mesh points to effectively encode motion information. To bridge the gap between Gaussian Splatting representation and mesh-based ground truth motion, we introduce a \emph{mesh-guided loss} that aligns the motion of Gaussian points with the corresponding mesh vertices. Our VAE is trained end-to-end with this mesh-guided loss and an image-level loss, enabling faithful compression of complex Gaussian Variation Fields. This approach reduces high-dimensional motion sequences to a compact 512-dimensional latent space, thus facilitating efficient diffusion modeling for 4D content generation.

Following the construction of our VAE, the 4D generative modeling naturally decomposes into canonical 3DGS generation and Gaussian Variation Field modeling. We leverage state-of-the-art 3D generative models~\cite{xiang2024structured} for the canonical component while focusing on modeling the Gaussian Variation Fields. To achieve this, we train a diffusion model to learn the latent space distribution of variation fields conditioned on the input video and canonical 3DGS, enabling controlled 4D content generation. Leveraging the compact nature of our latent space, we employ the Diffusion Transformer (DiT) architecture~\cite{peebles2023scalable}, augmented with temporal self-attention layers to capture smooth temporal dynamics across animations. The video frame features ~\cite{oquab2023dinov2} and the canonical 3DGS are taken as conditions for the diffusion model via cross-attention layers. Additionally, we incorporate positional priors into the diffusion model, enhancing its awareness of correspondences between canonical GS and their variation fields during the denoising process, thereby improving generation quality.

We train our model on a carefully curated diverse collection of animatable 3D objects from the Objaverse~\cite{deitke2023objaverse} and Objaverse-XL~\cite{deitke2023objaversexl}. Extensive evaluations demonstrate the superior video-to-4D generation quality of our method compared to existing approaches. Despite being trained on synthetic data, our model exhibits remarkable generalization capabilities when applied to in-the-wild video inputs, effectively creating impressive animations from in-the-wild animation sequences. We believe that our approach represents a notable step toward narrowing the gap between static 3D generation and 4D content creation, paving the way for generating high-quality 4D content.

\section{Related Work}

\noindent{\textbf{3D generation.}} Early GAN-based 3D generation approaches~\cite{wu2016learning, zhu2018visual, deng2021gram, chan2022efficient, gao2022get3d, skorokhodov20233d, zheng2022sdf, xiang2022gram} laid the foundation for 3D content synthesis, while diffusion-based methods~\cite{wang2023rodin, zhang2024rodinhd, zhang2024gaussiancube, tang2023volumediffusion, muller2023diffrf, he2024gvgen, cao2023large, shue20233d, jun2023shap, yariv2024mosaic, chen2023primdiffusion} advanced generation quality. Recent approaches have focused on latent space generation, either separating geometric modeling and appearance synthesis~\cite{zhao2024michelangelo,vahdat2022lion, zhang20233dshape2vecset, zhang2024clay, li2024craftsman, wu2024direct3d, zheng2023locally, ren2024xcube} or jointly modeling both~\cite{xiang2024structured, gupta20233dgen, xiong2024octfusion, jun2023shap, lan2024ln3diff, ntavelis2023autodecoding, yang2024atlas, chen20243dtopia}. Alternative methods~\cite{tang2023make, poole2022dreamfusion, sun2023dreamcraft3d, chen2023fantasia3d, liu2023zero, wang2023prolificdreamer} leverage pretrained 2D models~\cite{rombach2022high} through optimization techniques. Recent works~\cite{xiang2024structured,zhang2024clay} have achieved high-quality 3D asset generation with detailed geometry and appearance, establishing a foundation for 4D content creation.

\noindent{\textbf{4D reconstruction.}} Early 4D reconstruction methods~\cite{park2021nerfies,park2021hypernerf,fridovich2023k,pumarola2021d,cao2023hexplane} extended neural volumetric techniques for dynamic scenes, while recent Gaussian Splatting-based approaches~\cite{wu20244d,luiten2023dynamic,cotton2024dynamic,li2024gaussianbody,huang2024sc,kerbl20233d} offer improved efficiency. 
Typical 4D reconstruction methods often require significant optimization time per instance (\textit{e}.\textit{g}., 6 minutes for 4DGaussians~\cite{wu20244d} and over 30 minutes for K-planes~\cite{fridovich2023k}), making them impractical to use as a preliminary step in fitting 4D representations for generation.
In this paper, we explore an efficient approach to directly encode 4D mesh data for generative modeling in a single pass.

\begin{figure*}
    \centering
    \includegraphics[width=1.0\linewidth]{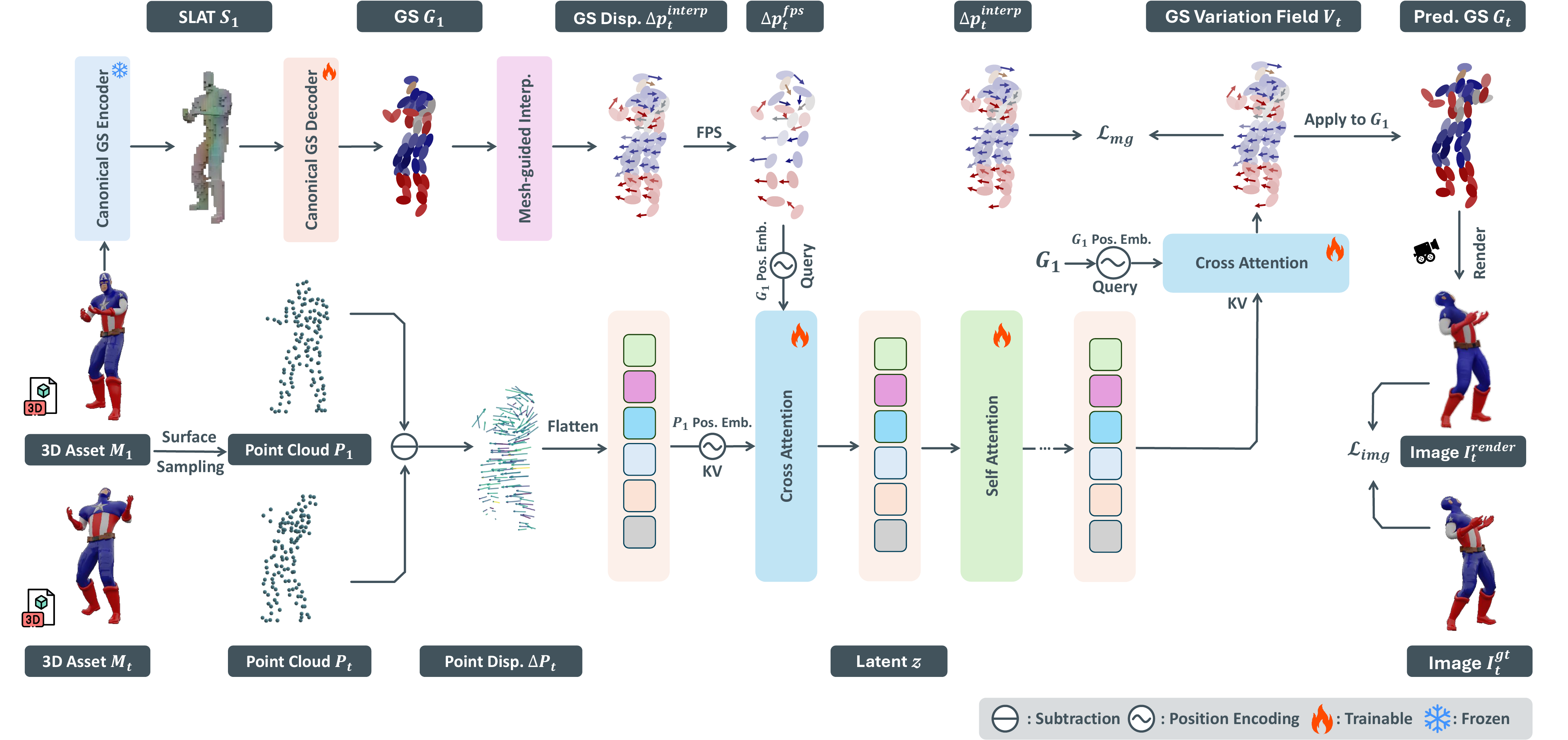}
    \caption{\textbf{Framework of 4DMesh-to-GS Variation Field VAE.} Our VAE directly encodes 3D animation data into Gaussian Variation Fields within a compact latent space, optimized through image-level reconstruction loss and the proposed \textbf{\emph{mesh-guided loss}}.} 
    \label{fig:enter-label}
\end{figure*}

\noindent{\textbf{Video-to-4D generation.}} Early attempts~\cite{jiang2023consistent4d, 
bahmani20244d, ren2023dreamgaussian4d, singer2023text} at video-to-4D generation 
predominantly relied on optimization-based approaches, utilizing pre-trained generative
priors~\cite{liu2023zero, shi2023zero123++, shi2023mvdream, wang2023imagedream} as guidance. These methods typically employ score distillation~\cite{poole2022dreamfusion} techniques to optimize either neural volumetric representations~\cite{mildenhall2021nerf} or 3D Gaussian Splatting~\cite{kerbl20233d}. These methods suffer from lengthy optimization times and SDS-related issues~\cite{wang2023prolificdreamer, liang2024luciddreamer} such as spatial-temporal inconsistency or poor input alignments. While some works~\cite{zeng2024stag4d,yin20234dgen} use pseudo-labels for better consistency, recent approaches~\cite{ren2025l4gm,xie2024sv4d,pan2024efficient4d,yang2024diffusion,zhang20254diffusion,liang2024diffusion4d} directly reconstruct 4D content from multiview images or videos. Notably, the introduction of large-scale 4D reconstruction models~\cite{ren2025l4gm} has significantly reduced the generation time from hours to seconds. However, most of these approaches often struggle to maintain consistent quality across temporal sequences due to inherent multiview inconsistency in 2D generation results.

\section{Method}

Given an input video sequence $\boldsymbol{\mathcal{I}} = \{I_t\}_{t=1}^T$ of an object, our goal is to generate a sequence of 3DGS models $\boldsymbol{\mathcal{G}} = \{G_t\}_{t=1}^T$ that captures both the shape, appearance, and motion of the object. We decompose this task into canonical GS $G_1$ creation (using the first frame as canonical) and Gaussian Variation Fields $\boldsymbol{\mathcal{V}} = \{\Delta G_t\}_{t=1}^T$ generation, where $\boldsymbol{\mathcal{V}}$ describes each Gaussian's attribute variations relative to $G_1$ over time. Our framework comprises two main components: (1) a \emph{direct 4DMesh-to-GS Variation Field VAE} that efficiently encodes 3D animation sequences into a compact latent space, and (2) a \emph{Gaussian variation field diffusion model} that learns the latent distribution of variation fields conditioned on the input video and canonical GS. The following sections detail each component.

\subsection{Direct 4DMesh-to-GS Variation Field VAE}
Extending 3DGS to generative modeling of dynamic content presents significant challenges. Fitting individual dynamic 3DGS representations for each animation instance is computationally expensive and scales poorly. Additionally, directly modeling temporal deformation of GS sequences with diffusion models is challenging due to the high dimensionality of both Gaussian quantities (\textit{e}.\textit{g}., typically over 100K in~\cite{kerbl20233d}) and the time dimension. Therefore, we propose an efficient autoencoding framework that directly encodes 3D animation data into Gaussian Variation Fields with a compact latent space, facilitating subsequent diffusion modeling.

\noindent{\textbf{Gaussian Variation Field encoding.}} Given a sequence of mesh animations $\boldsymbol{\mathcal{M}} = \{M_t\}_{t=1}^T$, we first convert them to point clouds $\boldsymbol{\mathcal{P}} = \{P_t | P_t\in \mathbb{R}^{N\times 3} \}_{t=1}^T$ through uniform surface sampling, where each point cloud contains $N$ points. The displacement fields $\{\Delta P_t | \Delta P_t \in \mathbb{R}^{N\times 3}\}_{t=1}^T$ are computed as temporal differences of corresponding points between frames:
\begin{equation}
    \Delta P_t = P_t - P_1,
\end{equation}
where $P_1$ is the canonical frame's point cloud. We then leverage a pretrained mesh-to-GS autoencoder $\mathcal{E}_{GS}$ and $\mathcal{D}_{GS}$ in~\cite{xiang2024structured} to obtain the canonical GS representation from canonical mesh $M_1$:
\begin{equation}
    \begin{split}
        S_1 = \mathcal{E}_{GS}(M_1),\\
        G_1 = \mathcal{D}_{GS}(S_1),
    \end{split}
\end{equation}
where $G_1 \in \mathbb{R}^{N_G\times 14}$ denotes the Gaussian parameters including positions $\bm{p}_1$, scales $\bm{s}_1$, rotation $\bm{q}_1$, colors $\bm{c}_1$, and opacity $\alpha_1$, with $N_G$ being the total number of canonical Gaussians. $S_1$ is the structured latent (\textsc{SLat}) representation for canonical GS (more details are included in the supplementary). We finetune $\mathcal{D}_{GS}$ to ensure coherent canonical GS reconstruction with their variation fields, while keeping $\mathcal{E}_{GS}$ frozen to leverage pretrained canonical GS diffusion models.

Inspired by 3DShape2VecSet~\cite{zhang20233dshape2vecset}, we employ a cross-attention layer to aggregate motion information from 3D animation sequences into a fixed-length latent representation. While directly using $G_1$ as query vectors is a straightforward approach, we find it leads to poor motion awareness. To enhance the network's sensitivity to mesh deformation, we introduce a \textbf{\textit{mesh-guided interpolation}} mechanism that generates motion-aware query vectors based on the spatial correspondence between $G_1$ and $P_1$.

Specifically, for each canonical Gaussian position $\bm{p}_1^i$, we identify its $K$ nearest neighbors in the canonical point cloud $P_1$ and compute their distances $\bm{d}_{i,k}$. To handle varying point densities across the mesh-sampled point cloud, we introduce an adaptive radius $r_i$ that adjusts the influence region based on the local point distribution. The interpolation weight $\bm{w}_{i,k}$ and adaptive radius $r_i$ are formulated as:
\begin{equation}
    \bm{w}_{i,k} = \exp(-\frac{\beta \bm{d}_{i,k}}{r_i^2}), \quad
    r_i = \sqrt{\frac{1}{K}\sum_{k=1}^K \bm{d}_{i,k}},
\end{equation}
where $\beta$ is a hyperparameter controlling the decay rate of interpolation weights with distance, with larger values producing more localized influence regions. We set $\beta=7.0$ in this paper.

We then interpolate the displacement fields $\Delta P_t$ for the $i$-th Gaussian at time $t$:
\begin{equation}
    \Delta \bm{p}^{interp}_{t,i} = \sum_{k=1}^K \frac{\bm{w}_{i,k}}{\sum_{k} \bm{w}_{i,k}} \Delta P_{t,n(i,k)}
\end{equation}
where $n(i,k)$ denotes the $k$-th nearest neighbor index. We perform farthest point sampling to $\Delta \bm{p}^{interp}_{t}$ based on their canonical positions to formulate our motion-aware query $\Delta \bm{p}^{fps}_{t} \in \mathbb{R}^{L\times 3}$ with reduced sequence length. The point cloud displacement fields $\Delta P_t$ serve as keys and values in the cross attention encoder. To preserve spatial relationships, we incorporate positional embedding $PE(\cdot)$ based on the canonical positions:
\begin{equation}
    \begin{split}
        Q_e &= f_{disp}(\Delta \bm{p}^{fps}_t) + PE(G_1), \\
        K_e &= V_e = f_{disp}(\Delta P_t) + PE(P_1), \\
        \bm{z} &= \text{CrossAttn}(Q_e, K_e, V_e),
    \end{split}
\end{equation}
where $f_{disp}$ is the displacement embedding layer. This process yields a latent representation $\bm{z} \in \mathbb{R}^{T \times L \times C}$, where $T$ is the number of temporal frames, $L$ is the latent size, and $C$ is the feature dimension. Notably, our encoding procedure compresses the sequence length from $N=8192$ to $L=512$, significantly reducing the subsequent diffusion modeling space. 


\begin{figure}
    \centering
    \includegraphics[width=0.95\linewidth]{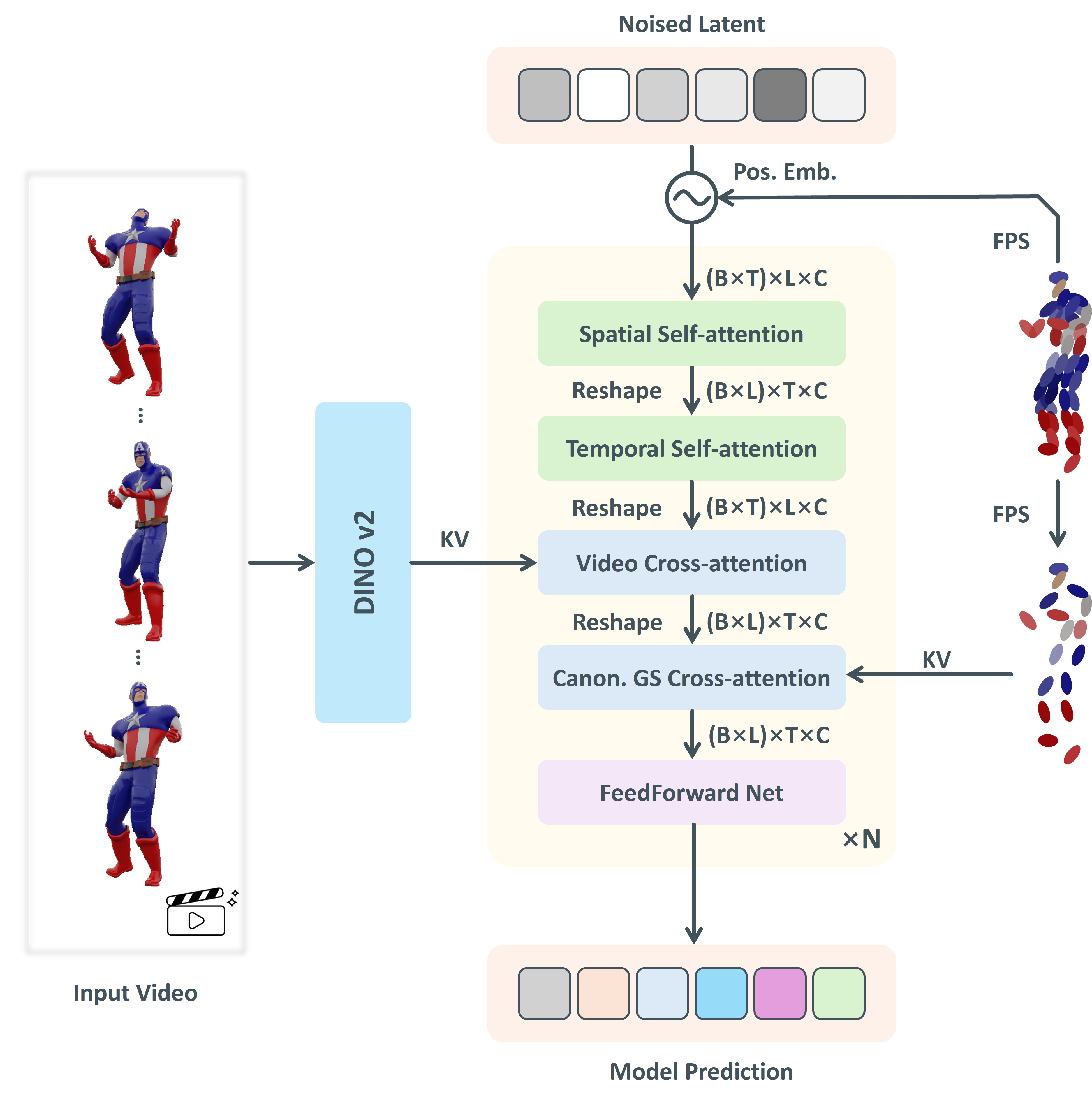}
    \caption{\textbf{Architecture of Gaussian Variation Field diffusion model.} Our model is built upon diffusion transformer, which takes noised latent as input and gradually denoises it conditioned on the video sequence and canonical GS.}
    \label{fig:dit}
\end{figure}

\noindent{\textbf{Gaussian Variation Field decoding.}} The decoding procedure first transforms the latent representation through $n$ layers of self-attention blocks to enable thorough motion information exchange. The decoder then maps this processed latent to a Gaussian Variation Field $\boldsymbol{\mathcal{V}}$, defined by the variations of Gaussian attributes $\Delta G_t=\{\Delta \bm{p}_t, \Delta \bm{s}_t, \Delta \bm{q}_t, \Delta \bm{c}_t, \Delta \alpha_t\}_{t=1}^T$. To ensure the decoder is aware of all canonical Gaussian attributes, we use all parameters of $G_1$ to query the latent output through a cross attention layer:
\begin{equation}
    \begin{split}
        Q_d = f_{gs}(G_1) &+ PE(G_1), \quad K_d = V_d = \bm{z}_n, \\
        \Delta G_t &= \text{CrossAttn}(Q_d, K_d, V_d),
    \end{split}
\end{equation}
where $f_{gs}$ is the embedding layer for the canonical Gaussians and $\bm{z}_n$ is the final self attention layer output. The final 3DGS sequence is obtained by:
\begin{equation}
    \boldsymbol{\mathcal{G}} = \{G_t\}_{t=1}^T = \{{G_1 + \Delta G_t}\}_{t=1}^T .
\end{equation}

\begin{table*}[t]  
	\centering  
	\small
    \caption{\textbf{Quantitative comparison of video-to-4D generation results.} Our method demonstrates consistent performance improvements across all metrics while maintaining efficient generation speed. The generation time is measured on a single A100 GPU.}
    \begin{tabular}{ccccccc}  
		\toprule
        \textbf{Method} & \textbf{PSNR$\uparrow$} & \textbf{LPIPS$\downarrow$} & \textbf{SSIM$\uparrow$} & \textbf{CLIP$\uparrow$} & \textbf{FVD$\downarrow$} & \textbf{Time$\downarrow$}  \\
		\midrule
        Consistent4D~\cite{jiang2023consistent4d} & 16.20 & 0.146 & 0.880 & 0.910 & 935.19 & $\sim$1.5 hr \\
        SC4D~\cite{wu2024sc4d} & 15.93 & 0.164 & 0.872 & 0.870 & 833.15 & $\sim$20 min \\
        STAG4D~\cite{zeng2024stag4d} & 16.85 & 0.144 & 0.887 & 0.893 & 1008.40 & $\sim$1 hr \\
        DreamGaussian4D~\cite{ren2023dreamgaussian4d} & 15.24 & 0.162 & 0.868 & 0.904 & 799.56 & $\sim$15 min\\
        L4GM~\cite{ren2025l4gm} & 17.03 & 0.128 & 0.891 & 0.930 & 529.10 & \textbf{3.5 s}\\
        \cellcolor{gray}{\textbf{Ours}} & \cellcolor{gray}{\textbf{18.47}} & \cellcolor{gray}{\textbf{0.114}} & \cellcolor{gray}{\textbf{0.901}} & \cellcolor{gray}{\textbf{0.935}} & \cellcolor{gray}{\textbf{476.83}} & \cellcolor{gray}{4.5s} \\
		\bottomrule  
	\end{tabular} 
	\label{tab:generation_res}  
\end{table*} 

\noindent{\textbf{Training objective.}} Our training objective consists of three main components. First, we employ image-level reconstruction loss between the rendered images $I_t^{render}$ from final predicted Gaussians and ground-truth images $I_t^{gt}$:
\begin{equation}
    \mathcal{L}_{img} = \mathcal{L}_{1} + \lambda_{lpips}\mathcal{L}_{lpips} + \lambda_{ssim}\mathcal{L}_{ssim},
\end{equation}
where $\lambda_{lpips}, \lambda_{ssim}$ are loss weights for perception loss~\cite{zhang2018unreasonable} and SSIM loss, respectively. To ensure faithful motion reconsturction, we introduce a \textbf{\textit{mesh-guided loss}} that aligns the predicted Gaussian displacements with pseudo ground-truth $\Delta p_{t}^{interp}$ obtained through \textbf{\textit{mesh-guided interpolation}}:
\begin{equation}
    \mathcal{L}_{mg} = \sum_{t=1}^T \| \Delta \mathbf{p}_t - \Delta \bm{p}_{t}^{interp} \|_2^2,
\end{equation}
which we find is crucial for motion reconstruction quality. Finally, to facilitate subsequent diffusion training, we also regularize the latent distribution with a KL divergence loss $\mathcal{L}_{kl}$. The total loss is: $\mathcal{L}_{total} = \mathcal{L}_{img} + \lambda_{mg}\mathcal{L}_{mg} + \lambda_{kl}\mathcal{L}_{kl}$, where $\lambda_{mg}, \lambda_{kl}$ are respective loss weights.

\subsection{Gaussian Variation Field Diffusion}

The diffusion process can be formalized as the inversion of a discrete-time Markov forward process. Let $\bm{z}^0 \in \mathbb{R}^{T\times L\times C}$ denote our initial latent of Gaussian Variation Field from the distribution $p(\bm{z})$. During the forward phase, we progressively corrupt this latent sequence by adding Gaussian noise over diffusion steps $s \in [0, S]$, following $\bm{z}^s := \alpha_s \bm{z}^0 + \sigma_s \bm{\epsilon}$, where $\bm{\epsilon} \sim \mathcal{N}(\mathbf{0}, \boldsymbol{I})$, and $\alpha_s, \sigma_s$ define the noise schedule. After sufficient diffusion steps, $\bm{z}^S$ approaches pure Gaussian noise. Generation is achieved by reversing this process, starting from random Gaussian noise $\bm{z}^S \sim \mathcal{N}(\mathbf{0}, \boldsymbol{I})$ and progressively denoising it to recover $\bm{z}^0$. 

The compact latent space enables us to build our diffusion model upon the powerful Diffusion Transformer (DiT) architecture~\cite{peebles2023scalable}. As illustrated in~\Cref{fig:dit}, the model takes noise-corrupted latent as input, and processes them through a series of transformer blocks for denoising. Each transformer block incorporates diffusion timestep information through adaptive layer normalization (adaLN) and a gating mechanism. Beyond the standard spatial self attention layers, we introduce dedicated temporal self-attention layers to ensure coherent motion generation across the sequence. 

To condition the generation process, we inject two types of features through cross-attention layers: (1) visual features $\boldsymbol{\mathcal{C}}^v = \{\bm{C}^v_t\}_{t=1}^T$ extracted from input video frames using DINOv2~\cite{oquab2023dinov2}, and (2) geometric features $\boldsymbol{\mathcal{C}}^{GS} = G_1^{fps}$ fartherest sampled from the static GS. We further incorporate positional embeddings based on canonical GS positions $\bm{p}_1^{fps}$ in our diffusion transformer, which strengthens the model's awareness of correspondences between canonical GS and their variation fields during the denoising process, thereby effectively improving the generation quality.

We parameterize our diffusion model $\hat{\bm{v}}_\theta$ to predict the velocity $v^s := \alpha_s \bm{\epsilon} - \sigma_s \bm{z}^0$ at each diffusion step $s$. The diffusion model is trained using:
\begin{equation}
    \mathcal{L}_{\text{simple}} = \mathbb{E}_{s, \bm{z}^0, \bm{\epsilon}}\left[\left\|\hat{\bm{v}}_\theta\left(\alpha_s \bm{z}^0 + \sigma_s \bm{\epsilon}, s, \boldsymbol{\mathcal{C}}\right) - \bm{v}^s\right\|_2^2\right],
\end{equation}
where $\boldsymbol{\mathcal{C}} = \{\boldsymbol{\mathcal{C}}^v, \boldsymbol{\mathcal{C}}^{GS}\}$ represents the conditional features of both $\boldsymbol{\mathcal{C}}^v$ and $\boldsymbol{\mathcal{C}}^{GS}$.

\begin{figure*}[t]
    \centering
    \includegraphics[width=1.0\linewidth]{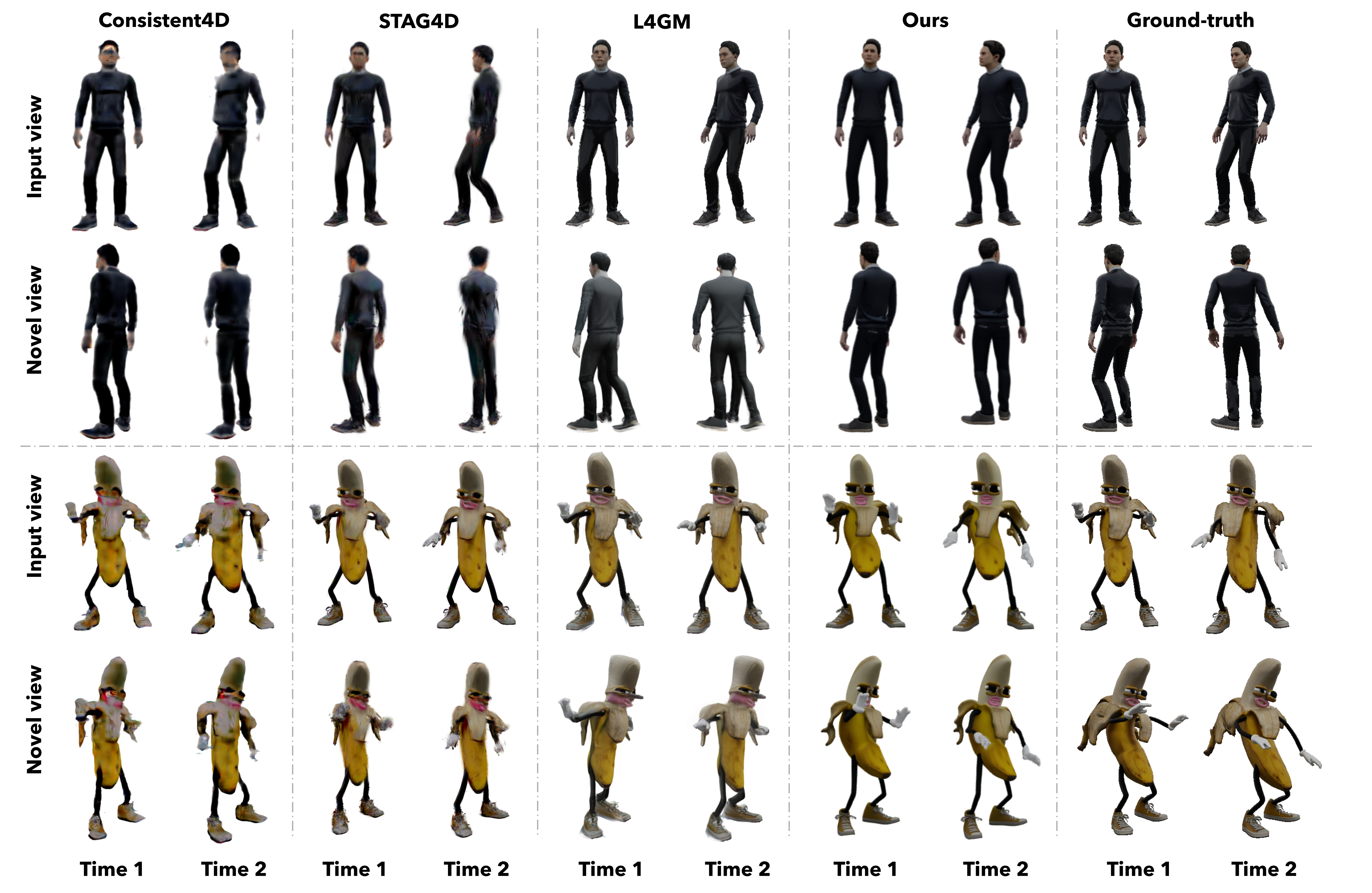}
    \caption{\textbf{Qualitative comparison with previous state-of-the-art methods.} Our model directly learns the distribution of Gaussian Variation Fields, enabling high-fidelity 4D generation with coherent temporal dynamics.}
    \label{fig:visual_comparison}
\end{figure*}

\subsection{Inference Pipeline}

During inference, our framework operates in a sequential pipeline. First, we obtain the canonical GS $G_1$ for the first frame using a pretrained 3D diffusion model~\cite{xiang2024structured}. Given an input video sequence $\{I_t\}_{t=1}^T$, we extract visual features and combine them with the farthest sampled canonical Gaussians as conditioning signals for our diffusion model. The diffusion model generates latent codes $\bm{z}$, which are subsequently decoded to obtain the Gaussian Variation Field $\boldsymbol{\mathcal{V}}$. The final animated Gaussian representation $G_t$ for each frame is obtained by applying these variations to the canonical Gaussians, effectively creating high-fidelity temporally coherent 4D animations.

\section{Experiments}

\subsection{Dataset and Metrics}

We conduct our experiments on Objaverse-V1 and Objaverse-XL~\cite{deitke2023objaverse}, following previous work in 4D content generation. After filtering for objects with high-quality animations, we utilize 34K objects for training. To evaluate the video-to-4D generation quality, we construct a comprehensive test set of 100 objects by combining 7 instances from the widely-used Consistent4D~\cite{jiang2023consistent4d} testset with 93 additional test instances from Objaverse-XL, ensuring a thorough evaluation with previous works. We render 4 novel views of each timestep for each instance. We assess the generation quality using multiple metrics: PSNR, LPIPS~\cite{zhang2018unreasonable}, and SSIM for frame-wise quality, and FVD~\cite{unterthiner2019fvd} for temporal consistency of the generated sequences. All evaluations are performed on renderings at $512 \times 512$ resolution.

\subsection{Implementation Details}

For our VAE implementation, the canonical Mesh-to-GS autoencoder is builds upon \textsc{Trellis}~\cite{xiang2024structured}, with training conducted in two stages: finetuning the sparse GS decoder $\mathcal{D}_{GS}$ on canonical 3D only for 150K iterations, followed by joint training with other modules for 200K iterations on 4D animation data. The VAE architecture employs point cloud size $N=8192$, latent size $L=512$, and feature dimension $C=16$. We optimize the VAE using AdamW with a learning rate $ 5e-6$ and $5e-5$ for $\mathcal{D}_{GS}$ and other modules respectively, using batch size 32. The diffusion model is trained on 24-frame sequences using AdamW optimizer with identical learning rate and batch size over 1300K iterations. We apply the cosine noise schedule~\cite{nichol2021improved} with 1000 timesteps for training the diffusion model. We set $T=24$ for training, and $T=32$ during inference to compare with prior works. For more implementation details, please refer to the supplementary materials.

\begin{figure*}[t]
    \centering
    \includegraphics[width=0.9\linewidth]{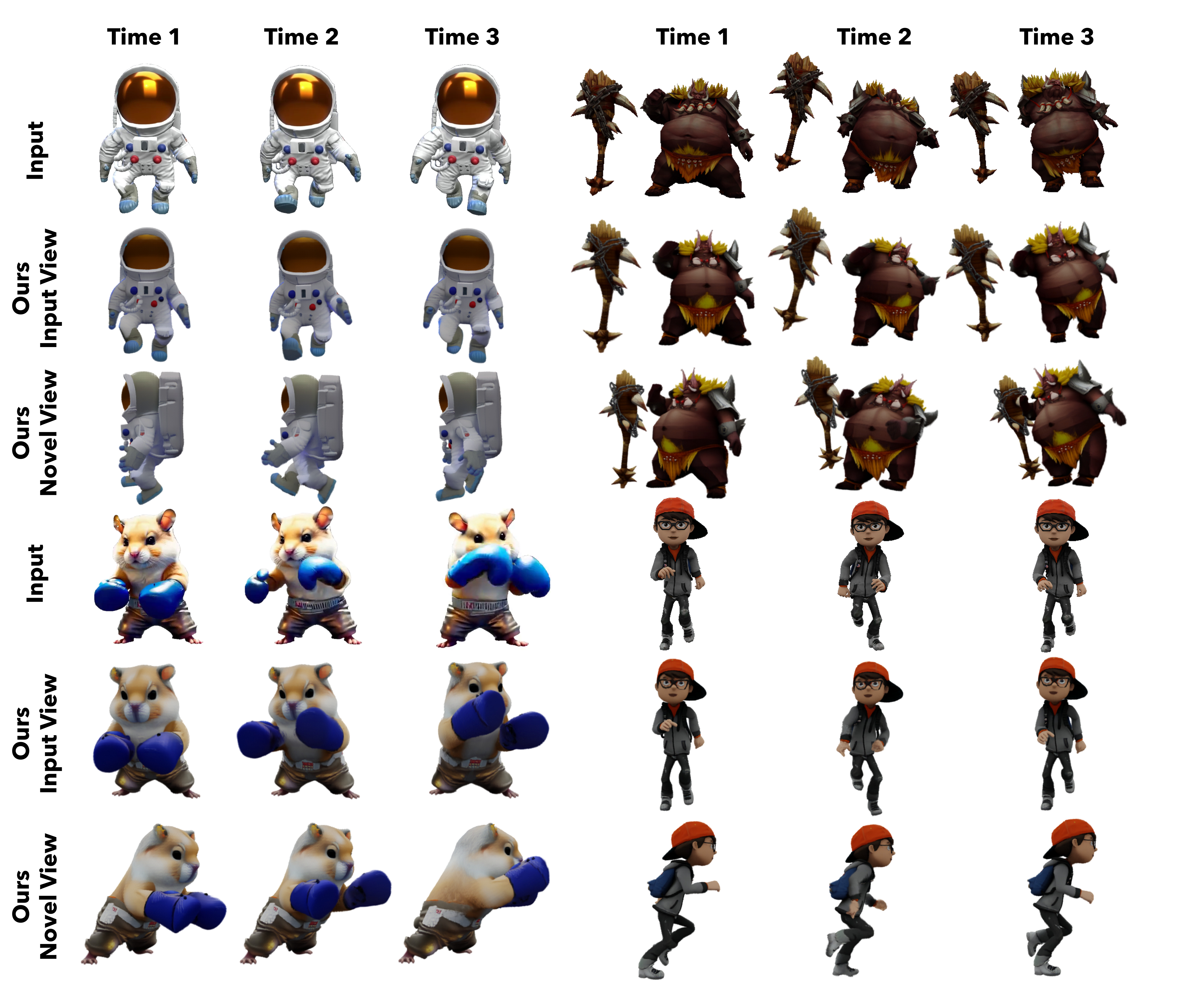}
    \caption{More generation results of our model including in-the-wild videos (left) and videos from test set (right).}
    \label{fig:more_visual_results}
\end{figure*}

\subsection{Main Results}

\noindent{\textbf{Quantitative comparisons.}} We compare the video-to-4D generation results of our model with previous state-of-the-art methods including both optimization-based approaches~\cite{jiang2023consistent4d, zeng2024stag4d, wu2024sc4d, ren2023dreamgaussian4d} and feedforward approach~\cite{ren2025l4gm}. As shown in~\Cref{tab:generation_res}, our method consistently outperforms existing approaches across all quality metrics, demonstrating both superior reconstruction fidelity and better temporal coherence. Unlike some prior works~\cite{jiang2023consistent4d, zeng2024stag4d, wu2024sc4d, ren2023dreamgaussian4d} require minutes to hours of optimization, our approach is also more efficient, taking 4.5 seconds to generate a 4D animation sequence (3.0 seconds for canonical GS creation and 1.5 seconds for Gaussian Variation Field diffusion), only slightly slower than feedforward reconstruction method L4GM~\cite{ren2025l4gm}. These quantitative results collectively validate both the effectiveness and efficiency of our proposed method.

\noindent{\textbf{Qualitative comparisons.}} We also provide qualitative comparisons with previous state-of-the-art methods in~\Cref{fig:visual_comparison}. The SDS-based approaches~\cite{jiang2023consistent4d,zeng2024stag4d} turn to generate results with blurry textures and poor geometry. The feedforward method L4GM leverages multiview images generated from 2D generative prior~\cite{shi2023mvdream} to reconstruct the 4DGS sequences. However, the results of L4GM suffer from 3D inconsistency of the generated multiview images. In contrast, our model directly generates the canonical GS and the Gaussian Variation Fields, capable of creating high-fidelity 3D consistent animations with coherent temporal dynamics. 

\noindent{\textbf{More visualization of generated results.}} \Cref{fig:more_visual_results} presents additional generation results from our method, including examples conditioned on both in-the-wild videos (left two cases) and test set videos (right two cases). Our model demonstrates high-quality generation capability with faithful motion reproduction. Despite being trained on synthetic data, the model exhibits strong generalization capability by effectively capturing motion patterns from in-the-wild video inputs. Furthermore, the model successfully handles challenging multi-object scenarios, highlighting the robustness of our approach.

\begin{table*}[t]  
	\centering  
	\small
    \caption{Ablation study of key factors in our VAE.}
    \vspace{-3mm}
    \begin{tabular}{ccccccc}  
		\toprule
        \textbf{Config.} & \textbf{Encoder Query Type} & \textbf{Mesh-guided Loss} & \textbf{Variation Attrs.} & \textbf{PSNR$\uparrow$} & \textbf{LPIPS$\downarrow$} & \textbf{SSIM$\uparrow$}  \\
		\midrule
         A. & $\bm{p}_t^{fps}$ & \xmark & $\Delta \bm{p}_t, \Delta \bm{s}_t, \Delta \bm{q}_t$ & 23.25 & 0.0678 & 0.936 \\
         B. & $\bm{p}_t^{fps}$ & \cmark & $\Delta \bm{p}_t, \Delta \bm{s}_t, \Delta \bm{q}_t$ & 26.17 & 0.0544 & 0.950 \\
         C. & $\Delta \bm{p}_t^{fps}$ & \cmark & $\Delta \bm{p}_t, \Delta \bm{s}_t, \Delta \bm{q}_t$ & 28.58 & 0.0478 & 0.958\\
         \cellcolor{gray}{D. (Ours)} & \cellcolor{gray}{$\Delta \bm{p}_t^{fps}$} & \cellcolor{gray}{\cmark} & \cellcolor{gray}{$\Delta \bm{p}_t, \Delta \bm{s}_t, \Delta \bm{q}_t, \Delta \bm{c}_t, \Delta \alpha_t$} & \cellcolor{gray}{\textbf{29.28}} & \cellcolor{gray}{\textbf{0.0439}} & \cellcolor{gray}{\textbf{0.964}} \\
		\bottomrule  
	\end{tabular} 
	\label{tab:vae_ablation}  
\end{table*} 

\begin{figure}
    \centering
    \includegraphics[width=1.0\linewidth]{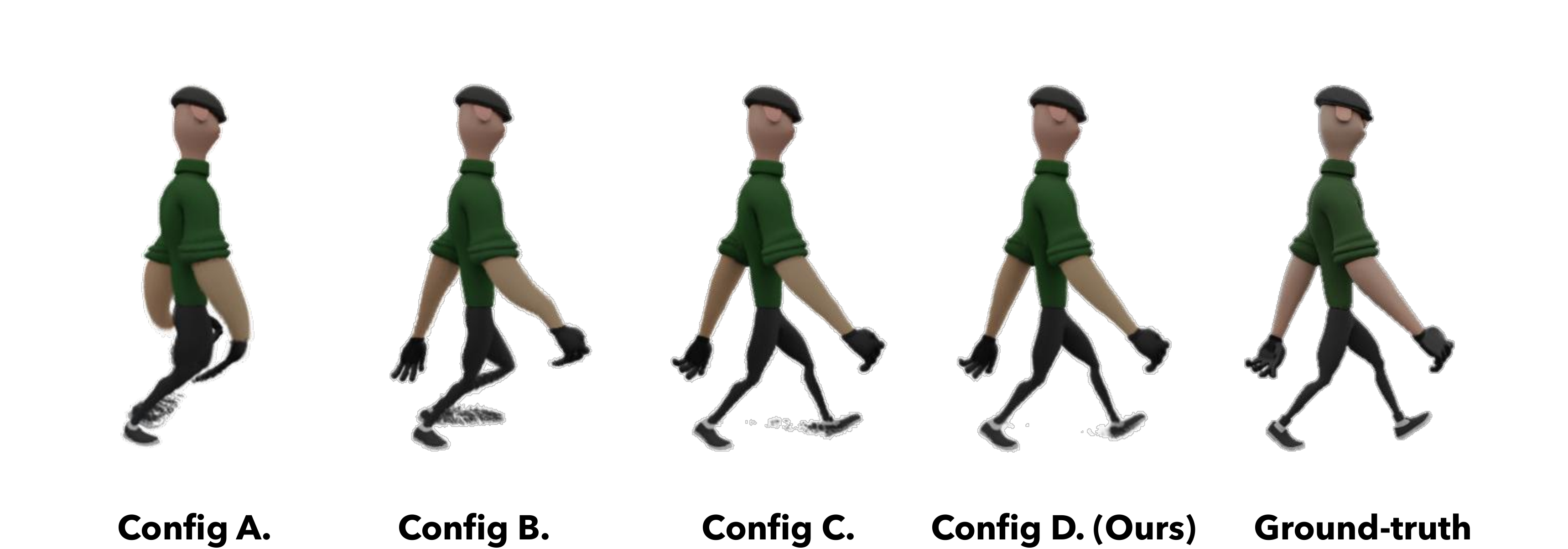}
    \caption{Visual ablation of VAE.}
    \label{fig:vae_ablation_fig}
\end{figure}

\begin{figure*}[t]
    \centering
    \includegraphics[width=0.8\linewidth]{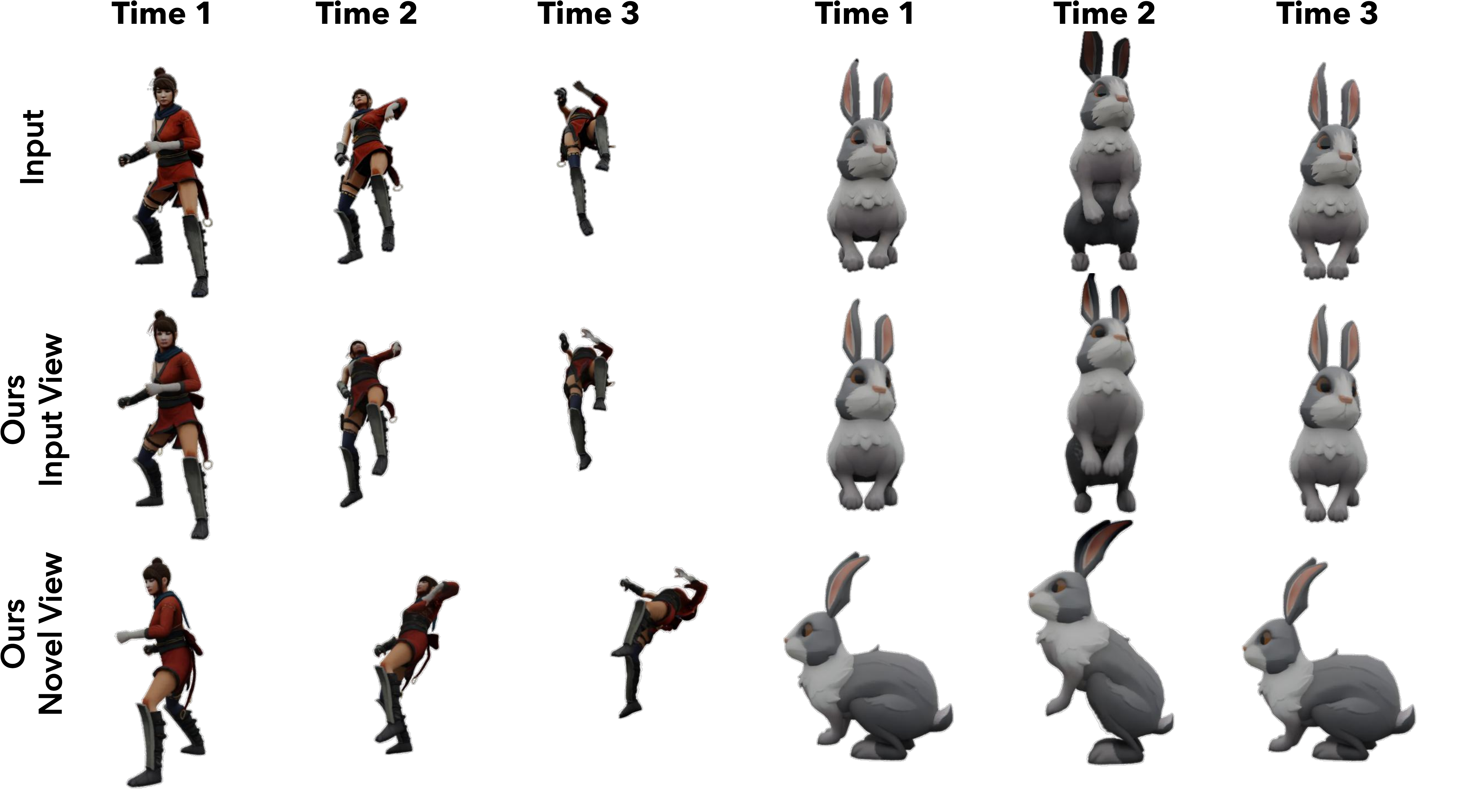}
    \vspace{-3mm}
    \caption{Our model is also capable of creating animations for existing 3D assets with conditional videos.}
    \label{fig:application_gt3dgs}
\end{figure*}

\begin{table}[t]  
	\centering  
	\small
    \caption{Ablation study of our diffusion model.}
    \setlength{\tabcolsep}{2pt}
    \begin{tabular}{ccccccc}  
		\toprule
        \textbf{Method} & \textbf{PSNR$\uparrow$} & \textbf{LPIPS$\downarrow$} & \textbf{SSIM$\uparrow$} & \textbf{CLIP$\uparrow$} & \textbf{FVD$\downarrow$}  \\
		\midrule
        Ours w/o Pos. Emb. & 17.86 & 0.121 & 0.897 & 0.931 & 547.20 \\
        \cellcolor{gray}{Ours} & \cellcolor{gray}{\textbf{18.47}} & \cellcolor{gray}{\textbf{0.114}} & \cellcolor{gray}{\textbf{0.901}} & \cellcolor{gray}{\textbf{0.935}} & \cellcolor{gray}{\textbf{476.83}} \\
		\bottomrule  
	\end{tabular} 
	\label{tab:diffusion_ablation}  
\end{table} 

\subsection{Ablation Study}

\noindent{\textbf{Ablation of our VAE.}} In~\Cref{tab:vae_ablation} and~\Cref{fig:vae_ablation_fig}, we analyze key components of our VAE. Our baseline (Config. A) starts with using positions of farthest sampled canonical GS $\bm{p}_t^{fps}$ as query for the encoder's cross-attention layer, with variation attributes limited to positions $\Delta \bm{p}_t$, scaling $\Delta \bm{s}_t$, and rotation $\Delta \bm{q}_t$ following previous 4DGS works~\cite{wu20244d}. Since we do not have ground-truth GS motion for explicit supervision, the VAE initially struggles with motion learning. After equipped with our \textbf{\emph{mesh-guided loss}}, the motion reconstruction capability is effectively improved through pseudo displacement supervision (Config. B). We then replace the encoding query with motion-aware $\Delta \bm{p}_t^{fps}$ using \textbf{\emph{mesh-guided interpolation}}, which successfully handles most of the motion sequences (Config. C). Finally, to give the model more flexibility to handle complex motion sequence, we incorporate color $\Delta \bm{c}_t$ and opacity $\Delta \alpha$ of Gaussian attributes to the variation fields, which further enhance the reconstruction capability of VAE.

\noindent{\textbf{Ablation of our diffusion model.}} We examine the importance of positional embeddings in our diffusion model training in~\Cref{tab:diffusion_ablation}. By incorporating positional prior based on canonical GS positions $\bm{p}_1^{fps}$, the diffusion transformer better captures the correspondence between spatial positions and their variations. Removing these positional embeddings leads to a significant performance drop, demonstrating their crucial role in achieving high-quality results.

\subsection{Application}
Despite being trained on single video inputs, our model is effective at animating existing 3D models according to the motions depicted in conditioned videos. More details are included in the supplementary materials. As demonstrated in~\Cref{fig:application_gt3dgs}, this approach produces high-quality animations that faithfully reproduce the target motions. Therefore, for real-world applications, the users can first generate 2D animations from the rendered images of their 3D models using off-the-shelf video diffusion models~\cite{openaisora,kong2024hunyuanvideo,kling,girdhar2024factorizing}, then employ our model to create corresponding 4D animations.

\section{Conclusion}
In this paper, we introduce a novel framework to address the challenging task of 4D generative modeling. To efficiently construct the large-scale training dataset and reduce the modeling difficulty for diffusion, we first introduce a \emph{Direct 4DMesh-to-GS Variation Field VAE}, which is able to efficiently compress complex motion information into a compact latent space without requiring costly per-instance fitting. Then, a \emph{Gaussian Variation Field diffusion} model that generates high-quality dynamic variation fields conditioned on input videos and canonical 3DGS. By decomposing 4D generation into canonical 3DGS generation and Gaussian Variation Field modeling, our method significantly reduces computational complexity while maintaining high fidelity. Quantitative and qualitative evaluations demonstrate that our approach consistently outperforms existing methods. Furthermore, our model exhibits remarkable generalization capability with in-the-wild video inputs, advancing the state of high-quality animated 3D content generation.

\noindent{\textbf{Acknowledgments.}} We extend our gratitude to all the reviewers for their constructive feedback. We also appreciate Jiaqi Lou for the assistance with chart refinement and the production of the supplementary video. 

{
    \small
    \bibliographystyle{ieeenat_fullname}
    \bibliography{main}
}
\clearpage
\appendix


\section{Additional Implementation Details}

\begin{table*}[t]  
	\centering 
	\small
        \caption{\textbf{Detailed configuration of model architecture.} \emph{SW} and \emph{FFN} denotes ``Shifted Window'' and ``FeedForward Net''. \emph{MSA}, \emph{MSSA}, \emph{MTSA}, \emph{MCA} stand for ``Multihead Self-Attention'', ``Multihead Spatial Self-Attention'', ``Multihead Temporal Self-Attention'' and ``Multihead Cross-Attention'', respectively.}
        \vspace{-3mm}
	\begin{tabular}{c|ccccccc}  
		\toprule
        \textbf{Network} & \textbf{\#Layer} & \textbf{\#Dim.} & \textbf{\#Head} & \textbf{Block Arch.} & \textbf{Special Modules} & \textbf{\#Param.} \\
        \midrule  
        $\boldsymbol{\mathcal{E}}_{GS}$ & 12 & 768 & 12 & 3D-SW-MSA + FFN & 3D Swin Attn. & 85.8M \\
        $\boldsymbol{\mathcal{D}}_{GS}$ & 12 & 768 & 12 & 3D-SW-MSA + FFN & 3D Swin Attn. & 85.1M \\
        VAE Transformer & 12 & 768 & 12 &  MSA / MCA + FFN & - & 125.21M \\
        Diffusion & 12 & 512 & 16 & MSSA + MTSA + MCA + FFN & QK Norm. & 105.51M \\
		\bottomrule
	\end{tabular}  
    \vspace{-3mm}
	\label{supp/tab:training_details}  
\end{table*}  

\subsection{Model Architecture}

We will detail the architecture of each model below, with the summary demonstrated in~\Cref{supp/tab:training_details}.
\subsubsection{Gaussian Variation Field Encoder}

Our encoder mainly comprises two parts: the canonical GS autoencoder $\mathcal{E}_{GS}$ and $\mathcal{D}_{GS}$ and a cross attention layer to create latent space for Gaussian Variation Fields. 

For the canonical GS autoencoder, we adopt the model architecture from~\cite{xiang2024structured}, which introduces a Structured Latent (\textsc{SLat}) representation for static 3D assets. This representation defines a set of local latents on a 3D grid, where each latent is associated with an active voxel intersecting with the surface of the 3D asset. The \textsc{SLat} representation effectively captures both the overall structure through active voxels and fine details through local latent codes. The canonical GS autoencoder is built using a transformer-based architecture. It first aggregates visual features from multiview images using a pre-trained DINOv2~\cite{oquab2023dinov2} encoder to create voxelized features. These features are then processed through a sparse transformer encoder that handles variable-length tokens corresponding to active voxels. The transformer incorporates shifted window attention in 3D space to enhance local information interaction while maintaining computational efficiency. The encoder outputs structured latents that follow a regularized distribution through KL-divergence penalties, which are then decoded to various representations. For this work, we only leverage its Gaussian representation decoder for our canonical GS autoencoding. $\mathcal{D}_{GS}$ is set to resolution 64, and decode to 8 Gaussians per voxel. We finetune the decoder $\mathcal{D}_{GS}$ while keeping the encoder $\mathcal{E}_{GS}$ frozen. 

For the cross attention layer, we adopt the vanilla full attention~\cite{vaswani2017attention} implementation. we set the motion-aware $\Delta \bm{p}^{fps}_{t} \in \mathbb{R}^{512 \times 3}$ using proposed \textbf{\textit{mesh-guided interpolation}} mechanism as query and point displacement fields $\Delta P_t \in \mathbb{R}^{8192 \times 3}$ from mesh as keys and values. Then the latent representation $\bm{z} \in \mathbb{R}^{512 \times 16}$ is obtained after the cross attention layer.

\subsubsection{Gaussian Variation Field Decoder}

For the Gaussian Variation Field decoder, we first adopt 12 layers of vanilla self attention for thorough information exchange. For the last cross attention layer to decode Gaussian Variation Fields $\Delta G_t \in \mathbb{R}^{N_G \times 14}$ The output feature of last self attention layer is set to keys and values, and we adopt all parameters of $G_1 \in \mathbb{R}^{N_G\times 14}$ as query, where $N_G$ is the total number of canonical GS. 

\subsubsection{Canonical GS Generation Model}

We adopt the model architecture from~\cite{xiang2024structured} to generate structure latent representation for further decoding to canonical GS, which follows a two-stage process. First, a structure generator creates the sparse structure by denoising a low-resolution feature grid using transformer blocks with adaptive layer normalization and cross-attention for condition injection. Then, a latent generator $\boldsymbol{\mathcal{G}}_{\mathrm{L}}$ generates local latents for the given structure using a sparse transformer architecture with downsampling and upsampling blocks. These two generators both adopt RMSNorm~\cite{zhang2019root} to the queries and keys (QK Norm.) in diffusion training. They are conditioned on image conditions through CLIP and DINOv2 features respectively, and are trained separately using a continuous flow matching objective. Since we freeze the $\mathcal{E}_{GS}$, we can directly leverage the pretrained image-to-3D model~\cite{xiang2024structured} to create canonical GS.

\subsubsection{Gaussian Variation Field Diffusion Model}

Our Gaussian Variation Field diffusion model builds upon the diffusion transformer architecture~\cite{peebles2023scalable}. To enable temporal coherence in generation, we introduce a temporal self-attention layer that complements the existing cross-attention, spatial self-attention, and feedforward layers. For video sequence conditioning, we extract frame-wise features using DINOv2~\cite{oquab2023dinov2} and incorporate the farthest-sampled canonical Gaussian Splatting to maintain awareness of the canonical 3D model. To enhance spatial consistency, we incorporate positional priors into the generation process. During training, we encode the Gaussian Variation Field latent along with their corresponding canonical GS positions to formulate positional embeddings. During inference, we directly utilize the positions from farthest-sampled Gaussian Splatting for positional embedding computation.

\subsection{Additional Training and Inference Details}

In this paper, we designate the first frame of each video as the canonical frame. For our Direct 4DMesh-to-GS Variation Field VAE training, we set the loss weights as follows: $\lambda_{lpips}=0.2$, $\lambda_{ssim}=0.2$, $\lambda_{mg}=1.0$, and $\lambda_{kl}=1e-6$. Computationally, the VAE training requires one day on 32 Nvidia Tesla V100 GPUs (32GB) for the first stage and two days on 8 Nvidia Tesla A100 GPUs (40GB) for joint training, while the diffusion model training spans approximately one week on 8 Nvidia Tesla A100 GPUs (80GB). During inference, we adopt the adaptive mode of DPM-Solver~\cite{lu2022dpm} with order 2, requiring approximately 18 steps per instance.

During inference, we address potential orientation misalignment between the generated canonical GS and input images through an azimuth alignment process similar to~\cite{ren2025l4gm}. Specifically, we render the canonical GS from multiple azimuth angles and compute image-level losses between these renders and the first video frame. We then transform the canonical GS according to the azimuth angle that yields the minimal loss, ensuring better alignment with the input video.

The in-the-wild conditional videos shown in the teaser (Figure 1 in main paper) are created by Kling~\cite{kling}. The walking astronaut and boxing rat video frames in Figure 5 of the main paper are sourced from consistent4D and Emu video~\cite{girdhar2024factorizing}, respectively.

\subsection{Additional Details of Creating Animation for Existing 3D Model}

To animate existing 3D models using our approach, users follow a simple pipeline: First, their 3D assets are rendered as multiview images. These images are then processed to extract and aggregate DINOv2 features. Using these features, we construct a canonical Gaussian Splatting representation through our $\mathcal{E}_{GS}$ encoder and $\mathcal{D}_{GS}$ decoder. Finally, animations are generated by our diffusion model, which takes both the canonical GS and a conditional video as input. Users can create these conditional videos using state-of-the-art video diffusion models~\cite{openaisora,kong2024hunyuanvideo,kling,girdhar2024factorizing} to specify their desired motion for the 3D model.

\section{Data Preparation Details}

Our training dataset consists of 34K 3D mesh animations sourced from Objaverse-V1~\cite{deitke2023objaverse} and Objaverse-XL~\cite{deitke2023objaversexl}. For Objaverse-V1, we utilize the curated set of 9K high-quality 3D animations from~\cite{liang2024diffusion4d}. For Objaverse-XL, we apply two filtering criteria: first, following~\cite{xiang2024structured}, we filter out samples whose average aesthetic score~\footnote{\href{https://github.com/christophschuhmann/improved-aesthetic-predictor}{https://github.com/christophschuhmann/improved-aesthetic-predictor}} across 4 rendered views of the first frame falls below 5.5; second, we remove sequences with minimal motion. This filtering process yields 25K additional animations from Objaverse-XL.

\section{Additional Ablation}

\noindent{\textbf{Ablation of $\mathcal{D}_{GS}$ Joint Finetuning.}} We investigate the importance of jointly finetuning the canonical GS decoder during our Direct 4DMesh-to-GS Variation Field VAE training. Starting from a pretrained canonical 3D $\mathcal{D}_{GS}$ checkpoint, we compare two settings: freezing $\mathcal{D}_{GS}$ while training other modules, and jointly training all modules (our approach). As shown in~\Cref{tab:supp_ablation}, joint training allows $\mathcal{D}_{GS}$ to receive feedback from animation reconstruction rather than being limited to static data only. This ensures the canonical GS reconstruction coherent with its corresponding variation fields.

\noindent{\textbf{Ablation of hyper-parameters in mesh-guided interpolation.}} We ablate the hyper-parameters including nearest neighbors $K$, and distance decay rate $\beta$ of interpolation in~\Cref{tab:supp_hyperparameter_ablation}. Our setting ($K=8, \beta=7.0$) yields optimal results. Performance is relatively stable for other values, showing reasonable robustness.

\begin{table}[t]  
	\centering  
	\small
    \caption{Additional ablation of our proposed VAE.}
    \begin{tabular}{ccccccc}
		\toprule
        \textbf{Model} & \textbf{PSNR$\uparrow$} & \textbf{LPIPS$\downarrow$} & \textbf{SSIM$\uparrow$}  \\
		\midrule
        Ours w/o $\mathcal{D}_{GS}$ Finetuning & 28.80 & 0.0460 & 0.962 \\
        \cellcolor{gray}{Ours} & \cellcolor{gray}{\textbf{29.28}} & \cellcolor{gray}{\textbf{0.0439}} & \cellcolor{gray}{\textbf{0.964}} \\
		\bottomrule  
    \end{tabular} 
	\label{tab:supp_ablation}  
\end{table} 

\begin{table}[t]  
	\centering  
	\small
    \caption{Additional ablation of hyper-parameters in our mesh-guided interpolation.}
    \setlength{\tabcolsep}{1pt}
    \begin{tabular}{ccccc|ccccc}  
		\toprule
        $K$ & $\beta$ & \textbf{PSNR$\uparrow$} & \textbf{LPIPS$\downarrow$} & \textbf{SSIM$\uparrow$} & $K$ & $\beta$ & \textbf{PSNR$\uparrow$} & \textbf{LPIPS$\downarrow$} & \textbf{SSIM$\uparrow$} \\
		\midrule
         16 & 7.0 & 28.38 & 0.0464 & 0.960 & 8 & 10.0 & 28.55 & 0.0462 & 0.961\\
         \cellcolor{gray}{\textbf{8}} & \cellcolor{gray}{\textbf{7.0}} & \cellcolor{gray}{\textbf{29.28}} & \cellcolor{gray}{\textbf{0.0439}} & \cellcolor{gray}{\textbf{0.964}} & \cellcolor{gray}{\textbf{8}} & \cellcolor{gray}{\textbf{7.0}} & \cellcolor{gray}{\textbf{29.28}} & \cellcolor{gray}{\textbf{0.0439}} & \cellcolor{gray}{\textbf{0.964}} \\
         4 & 7.0 & 28.94 & 0.0451 & 0.963 & 8 & 4.0 & 29.04 & 0.0446 & 0.963 \\
         1 & 7.0 & 28.22 & 0.0465 & 0.960 & 8 & 1.0 & 28.64 & 0.0457 & 0.962 \\
		\bottomrule  
	\end{tabular} 
	\label{tab:supp_hyperparameter_ablation}  
\end{table} 

\begin{figure}[t]
    \centering
    \includegraphics[width=0.95\linewidth]{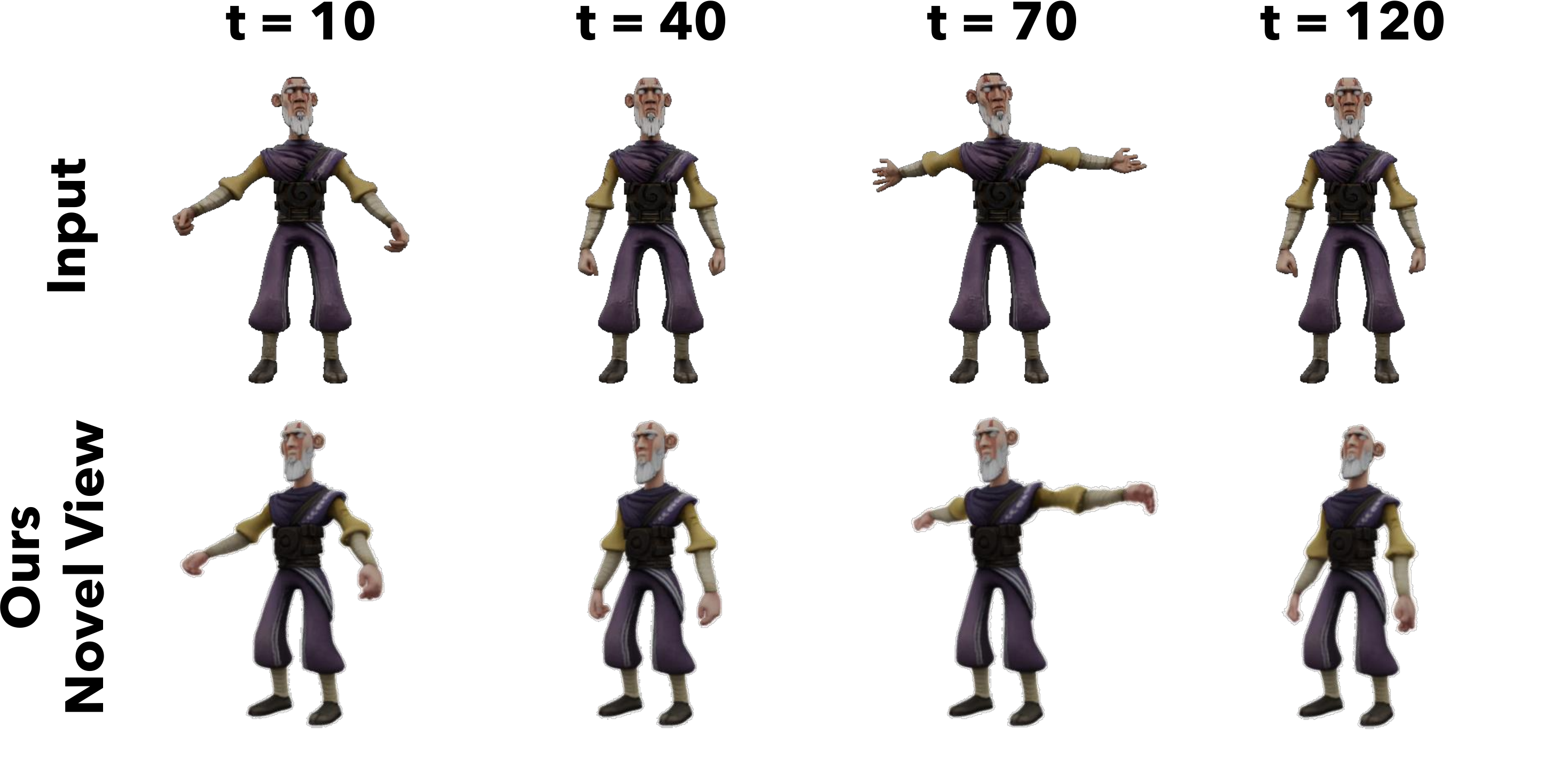}
    \caption{Sample of our autoregressive generation result.}
    \label{fig:supp-ar}
\end{figure}

\section{More Results}

\subsection{Autoregressive Generation Results for Temporal Generalization}
Temporal generalization is a known challenge in 2D/3D video generation. In our case, we can employ an autoregressive approach during inference for videos exceeding our training length: the GS from the last frame of a generated segment serves as the canonical GS for inferring the next segment's variation fields, which allows for coherent long animations. We show a 120-frame generated sample using such an approach in~\Cref{fig:supp-ar}.

\subsection{VAE Reconstruction Results}
As illustrated in~\Cref{fig:supp_vae}, we demonstrate the reconstruction capabilities of our proposed Direct 4DMesh-to-GS Variation Field VAE. Our method efficiently encodes both canonical GS and their temporal variations from 4D meshes in a single pass, eliminating the need for time-consuming per-instance fitting procedures. The results demonstrate our model's effectiveness in preserving both geometric fidelity and motion dynamics.

\begin{figure}[t]
    \centering
    \includegraphics[width=1.0\linewidth]{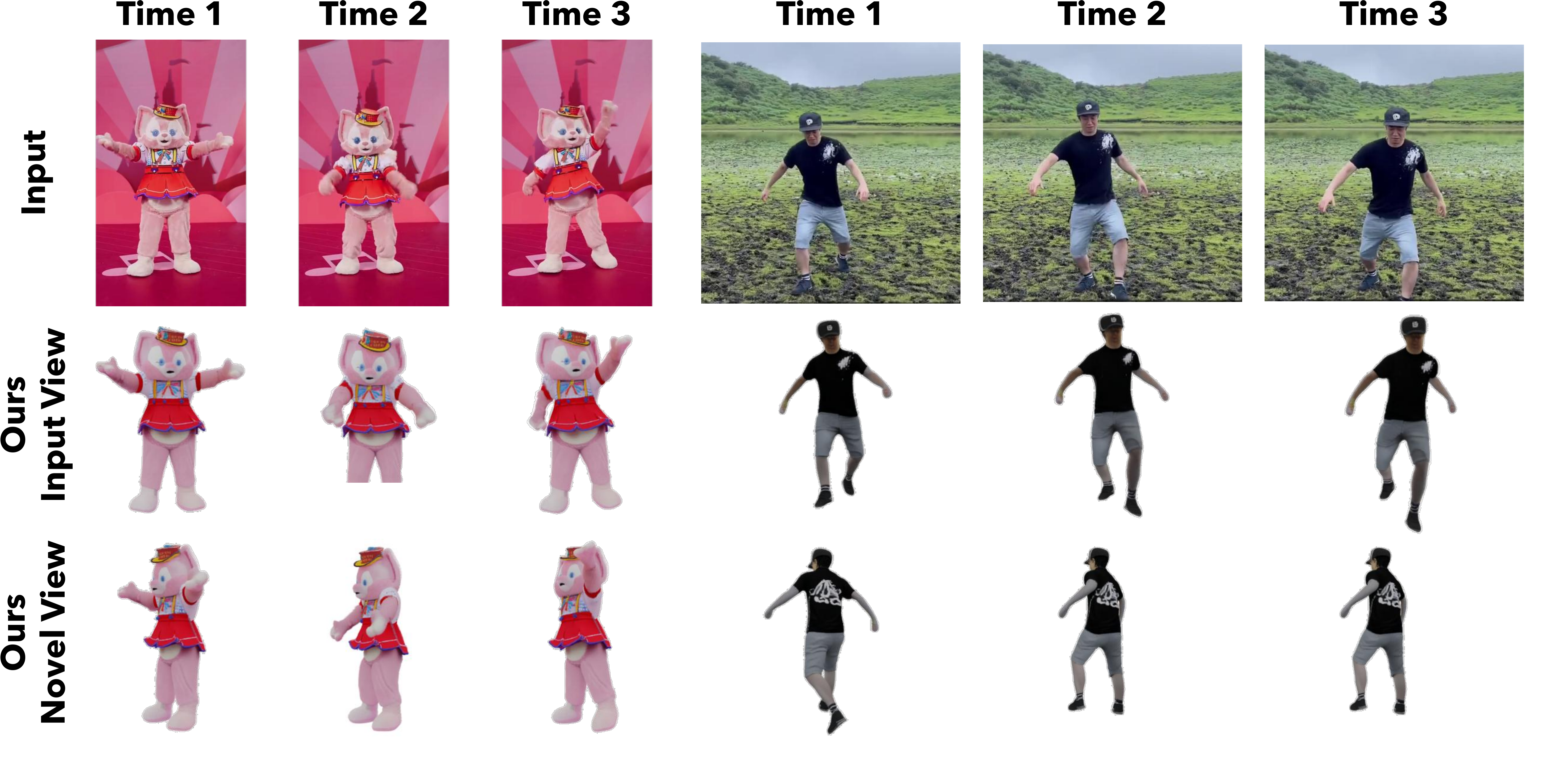}
    \caption{More generation results of real-world input videos.}
    \label{fig:supp-real-world}
\end{figure}

\subsection{More Visual Comparison with SOTA Methods}
As illustrated in~\Cref{fig:supp_vc}, we provide extensive visual comparisons with state-of-the-art methods. Our approach demonstrates consistent superiority across diverse test cases, achieving better results in terms of both visual fidelity and temporal motion coherence.

\subsection{More Reults of Animating Existing 3D Models}
As shown in~\Cref{fig:supp_gt3dgs}, we demonstrate additional results showcasing our method's capability to animate existing 3D models using conditional videos. Our approach successfully extracts and transfers motion patterns from the input videos, generating high-fidelity animations that faithfully preserve both geometric and temporal characteristics.

\subsection{Additional Results on Real-World Video Inputs}
Although our model is trained on synthetic data, it effectively generalizes to real-world video inputs. \Cref{fig:supp-real-world} presents additional results, demonstrating the model's robust generalization capabilities.

\section{Borader Impact}
Like all generative models, our video-to-4D generation framework requires careful consideration of societal implications. While we mitigate certain ethical concerns by training exclusively on synthetic 3D animations from the Objaverse dataset, thus avoiding privacy and copyright issues associated with real-world data, we acknowledge potential risks. The ability to generate animated 3D content from videos could be misused for creating misleading content. We therefore emphasize the importance of establishing clear guidelines for the responsible deployment of video-to-4D generation technology.

\section{Limitation Discussion and Future Work}

\begin{figure}[t]
    \centering
    \scriptsize
    \begin{tabular}{cc}
        \includegraphics[width=0.4\linewidth]{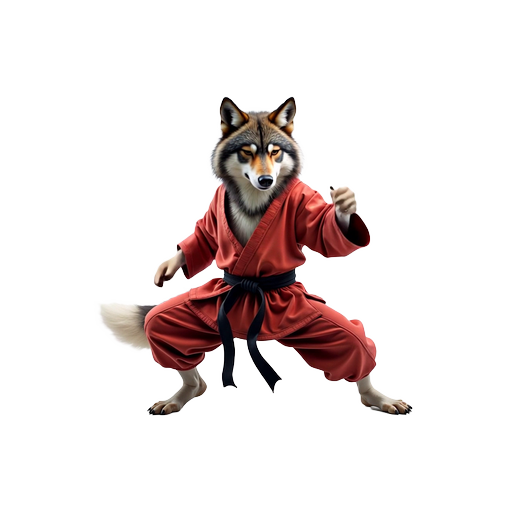} & \includegraphics[width=0.4\linewidth]{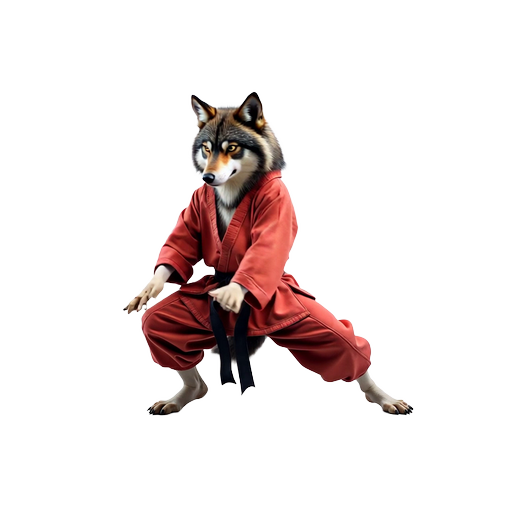} \\ 
        \makecell[c]{Video condition \\ frame 0} & \makecell[c]{Video condition \\ frame 1} \\
        \includegraphics[width=0.4\linewidth]{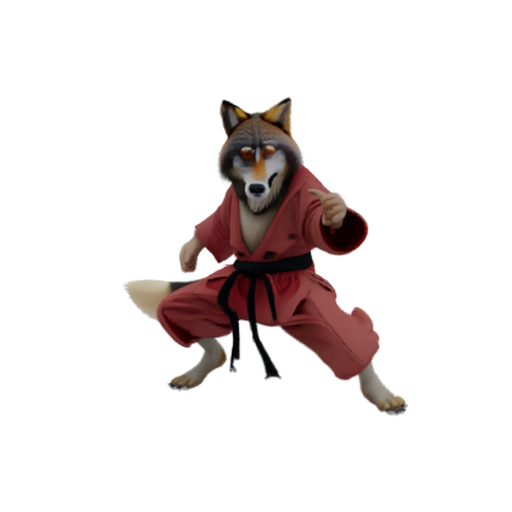} & \includegraphics[width=0.4\linewidth]{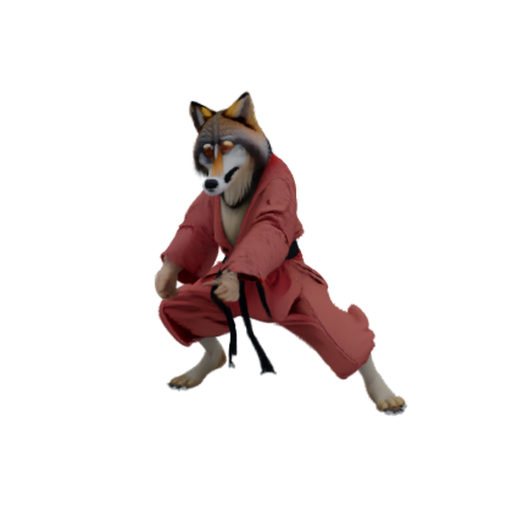} \\
        \makecell[c]{Canonical GS \\ frame 0} & \makecell[c]{Our Animation \\ frame 1} 
    \end{tabular}
    \caption{\textbf{Failure case.} When the pretrained static 3D generative model produces canonical GS that are not well-aligned with the conditional video frames, our Gaussian Variation Field diffusion model struggles to bridge this inconsistency, resulting in suboptimal animations.}
    \label{fig:supp_limitation}
\end{figure}

While our model demonstrates impressive results in video-to-4D generation, it has certain limitations. Our two-stage generation process first employs a pretrained static 3D generative model to create canonical Gaussian Splatting representations, which then serve as conditions for our diffusion model to generate Gaussian Variation Fields. A notable limitation arises when the static 3D generative model~\cite{xiang2024structured} produces canonical GS that exhibits discrepancies with the conditional video, such as mismatched head pose, incorrect eyes or light effects in~\Cref{fig:supp_limitation}, potentially creating inconsistencies in the final animation. To address this limitation, future work could explore either fine-tuning the static 3D model to ensure better image alignment or developing an end-to-end 4D diffusion framework that jointly generates both the canonical representation and its temporal variations.

\begin{figure*}[t]
    \centering
    \includegraphics[width=1.0\linewidth]{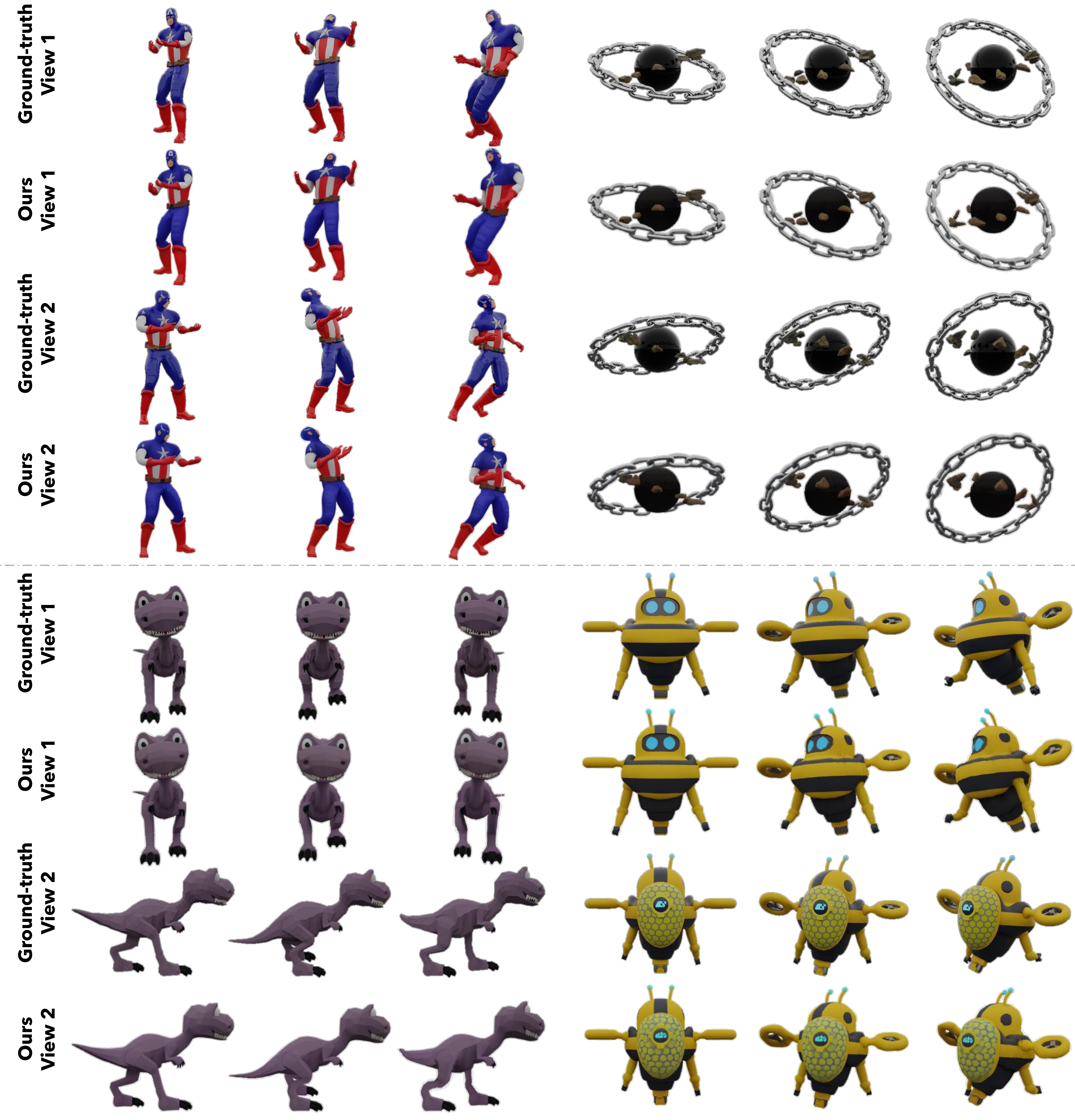}
    \caption{Additional visual results of VAE reconstruction.}
    \label{fig:supp_vae}
\end{figure*}

\begin{figure*}[t]
    \centering
    \includegraphics[width=0.95\linewidth]{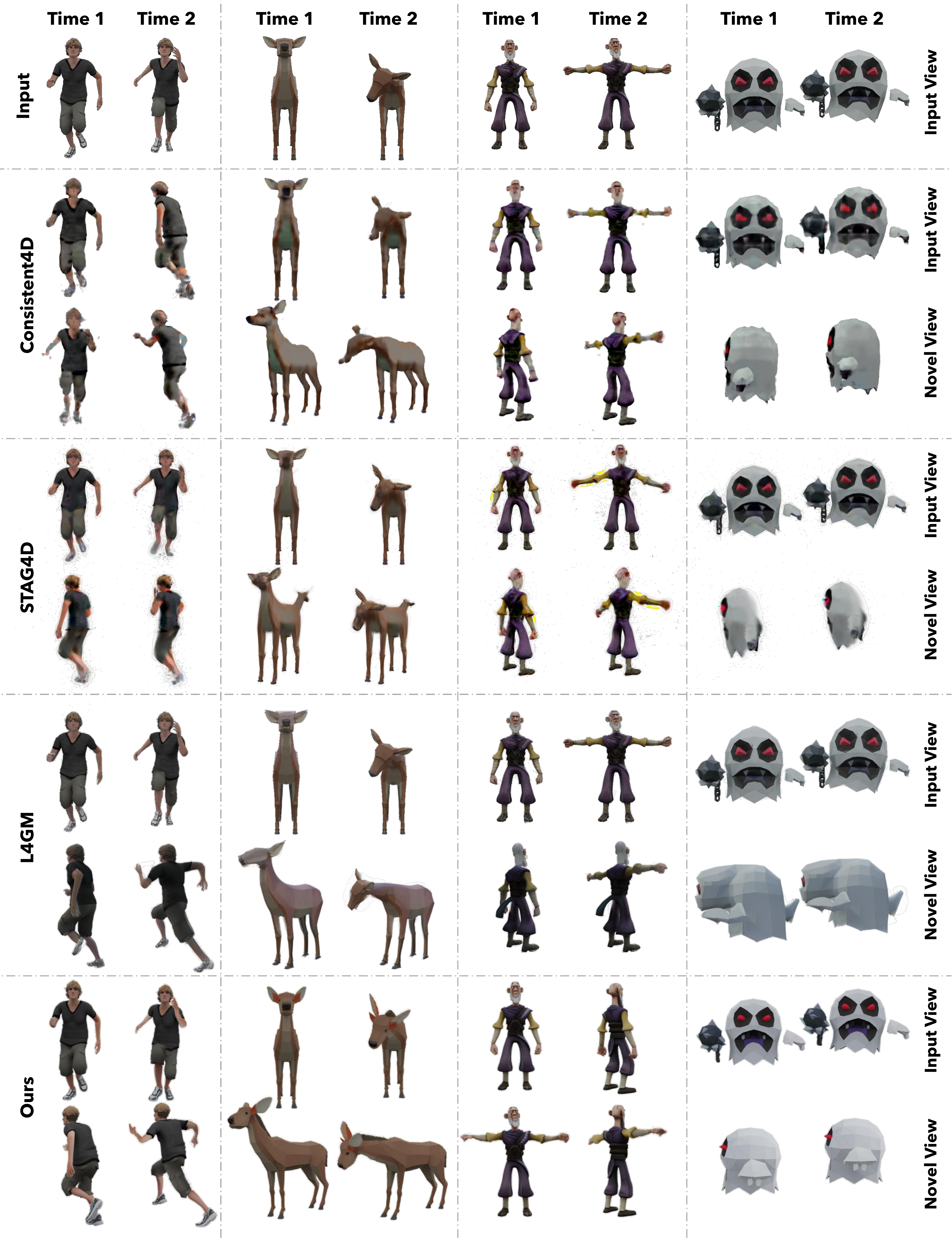}
    \vspace{-3mm}
    \caption{More visual comparison with SOTA Methods.}
    \label{fig:supp_vc}
\end{figure*}

\begin{figure*}[t]
    \centering
    \includegraphics[width=1.0\linewidth]{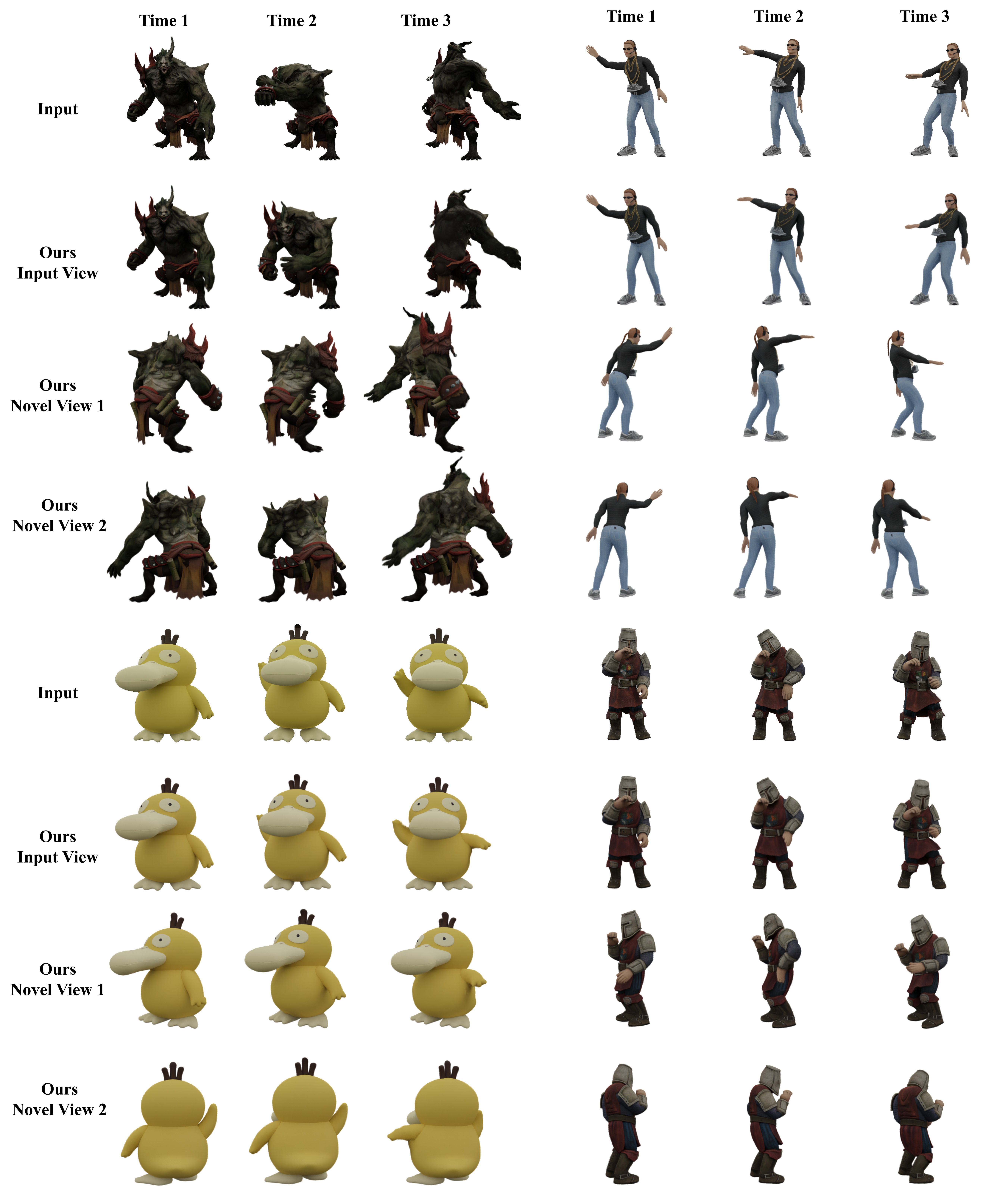}
    \caption{More results of animating existing 3D model input with conditional videos.}
    \label{fig:supp_gt3dgs}
\end{figure*}

\end{document}